\documentclass[conference]{IEEEtran}
\usepackage{cite}
\ifCLASSINFOpdf
  \usepackage[pdftex]{graphicx}
  \DeclareGraphicsExtensions{.pdf,.jpeg,.png}
\fi

\usepackage[cmex10]{amsmath}
\usepackage{array}

\begin{document}
\title{Grounded Lexicon Acquisition - Case Studies in Spatial Language}

\author{\IEEEauthorblockN{Michael Spranger}
\IEEEauthorblockA{Sony CSL\\
6, rue Amyot\\
75005 Paris, France\\
Email: spranger@csl.sony.fr}
}

\maketitle

\begin{abstract}
This paper discusses grounded acquisition experiments of 
increasing complexity. Humanoid robots acquire English spatial lexicons 
from robot tutors. We identify how various spatial 
language systems, such as projective, absolute and 
proximal can be learned. The proposed learning mechanisms
do not rely on direct meaning transfer or direct access to world 
models of interlocutors. Finally, we show how multiple systems
can be acquired at the same time.
\end{abstract}
\IEEEpeerreviewmaketitle

\section{Introduction}
There is an ongoing debate how conceptual development and
linguistic development interact. On the one hand, there is the idea
that many parts of the human conceptual repertoire are 
pre-determined by biological constraints. While certainly, biological constraints 
play an important role for linguistic development both ontogenetically and
phylogenetically, recent evidence points to considerable flexibility on the
conceptual and linguistic level \cite{evans2009universals}. The evidence
for this comes from studies in linguistic diversity, which show tremendous
cross-linguistic variation on the syntactic, semantic and conceptual level \cite{levinson2003space,svorou1994grammars}.
For instance, while English has an elaborate system of projective
categories such as ``front'' and ``back'', other languages such as Tzeltal 
\cite{brown2008up} rely on absolute geocentric features in the environment to
 build conceptual spaces.

Lexicon acquisition by artificial systems is an important topic and has been 
treated in a number of studies \cite{bailey1997modeling,roy2002learning,vogt2003strategies,steels2002aibo}.
Also in the ICDL/Epirob community this is a recurring theme that has received considerable
attention (e.g. see \cite{baxter2012seasnake,saunders2012robot} for recent examples).
However, many of these studies are done in pure simulation, or presuppose shared meaning spaces 
and/or shared world models. Another simplification often made is to use discrete or 
discretised meaning spaces. This paper tries to remedy this situation and explores what are the necessary
computational mechanisms allowing artificial agents to acquire grounded spatial lexicon systems
that can be used both in language understanding and language production.
Learners directly operate on continuous perceptual spaces and conceptual development
is directly organising the sensorimotor space.

We setup experiments with humanoid robots in which a tutor robot
is teaching a learner robot spatial relations in progressively more and
more complex tasks. We start by exploring the acquisition of single spatial
relation systems (e.g. only proximal relations such as ``near'' and ``far''). Subsequently,
we move to the simultaneous acquisition of different category systems. Finally, we discuss
systems which are flexible enough to acquire any kind of category system without a priori 
assumptions about the systems themselves.

This paper is part of a larger research effort that tries to understand the
basic processing principles \cite{spranger2011german}, acquisition and 
evolution of spatial language \cite{spranger2012grammar} with a particular 
focus on linguistic variation. Here we focus on lexicon 
acquisition. We exemplarily carry out acquisition experiments for English,
which features different spatial relation systems that represent rather completely
the different types of systems we find in different combinations also in other languages.

\section{Experimental Setup}
\begin{figure}
\centering
\includegraphics[width=0.9\columnwidth]{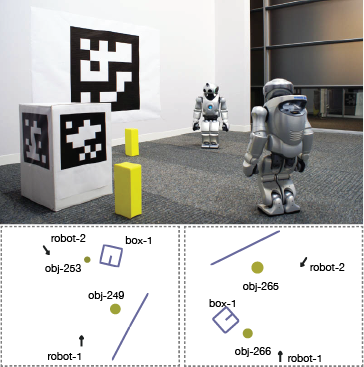}
\caption{Spatial language game setup. The world models computed 
by each robot are shown left and right. Each robot estimates the
position of objects from his own perspective. The arrows 
signify position and orientation of the robots. Each robot is himself at the
center of the coordinate system (robot-1). The blue line
in each world model represents the global direction marker on the wall. Yellow
circles represent colour and position of the bricks in the scene. The blue
rectangle shows position and orientation of the box (not important for this paper). }
\label{f:setup}
\end{figure}

We use \emph{language games} \cite{steels2001language} -- 
routinised interactions between communication partners. 
In our case, Sony humanoid robots 
interact in an office environment. Always two robots interact. One is a 
tutor agent, the other is a learner. Before an interaction, robots are scanning the 
environment for objects. Each robot's vision system singles
out objects such as coloured blocks, marker-augmented boxes and wall-pasted markers 
from the environment and estimates object properties such as distance, angle and 
color \cite{spranger2012perception}. Figure \ref{f:setup} shows the experimental setup 
and the objects involved. 

An interaction begins with random assignment of roles to the learner and the
tutor. One is the speaker, the other acts as the hearer. The interaction
then proceeds with the following steps.
\begin{enumerate}
\item The speaker selects one object from his world model, further called 
the topic $t$. For the purpose of this paper only blocks are chosen (yellow objects
in Figure \ref{f:setup}).
\item The speaker tries to find a spatial relation for describing the topic. 
\item The speaker looks up the word associated with this spatial relation 
in memory and produces the word. 
\item The hearer looks up which relation is associated with this word in memory
and examines his world model to find out whether there is a unique object 
which satisfies this relation. 
\item The hearer then points to this object in the world.
\item The speaker checks whether the hearer selected the same 
object as the one the speaker originally chose. If they are the same, 
the game is a success and the 
speaker signals this outcome to the hearer. 
\item If the game is a failure, the speaker points to the topic. 
\end{enumerate}

This script explicitly defines the feedback and input the learner 
deals with. He gets linguistic feedback (if he is hearer),
positive or negative feedback and pointing. Never is there any
direct meaning transfer or world model sharing between the tutor
and the learner. In fact, the world models of tutor and learner are always
necessarily different because the world is perceived by each
agent separately \cite{spranger2012deviation}.

Importantly, tutor and learner take on roles as speaker and hearer randomly. 
This means that learners immediately start trying to speak themselves even though the
command of the language they are learning might be rudimentary. Consequently, interactions 
can fail in various ways. For instance, the learner (as a speaker) might be unable to find 
a spatial relation for discriminating the topic (step 2), or
the learner (as a hearer) might not know the spatial term (step 4). Moreover, the hearer 
(tutor or learner) might point to the wrong object (step 5). 

\section{Representing Spatial Categories}
The English locative system can be broadly categorised into three different 
classes of categories . 

\paragraph{Proximal categories} such as ``near'' and ``far'' rely on proximity to
some particular landmark object.
\paragraph{Projective categories} are categories such as ``front'', ``back'', ``left'' and ``right''.
These categories are primarily angular categories signifying a direction. The landmark
object provides a special direction ``the front''. All other
categories follow from this pivot direction. The projective system mostly relies on object
features to determine which is the front. This is called intrinsic frame of reference
in the literature (we are ignoring the relative frame of reference in this paper).
In this experiment the robots use their own front for determining the pivot direction.
\paragraph{Absolute categories} such as ``north'', ``south'', ``east'' and ``west'' 
rely on a compass directions,  with the pivot direction to the magnetic north pole. 
The absolute system relies on features of the environment to determine the 
layout of the angles. In other languages, other geocentric features of the environment
are used. For example, in Tenejapan the directions uphill/downhill are used \cite{brown2008up}.
In the experiments discussed here, the wall marker is used as a global
direction on the scene.

We represent spatial categories using a either prototypical angle (absolute, projective) or 
distance (proximal). Additionally, each category is parameterised by a $\sigma$
value. The two parameters to a category, prototype value and $\sigma$, describe 
the similarity function of the category. 
\begin{eqnarray}
\label{e:angular-category-similarity}
\operatorname{sim}_{a}(o,c)&:=&e^{-\frac{1}{2 \sigma_c} d_a(o,c)}
\end{eqnarray}
where $o$ is some object, $c$ a category and $d_a$ is a distance 
function defined between prototypes and objects. Distance functions 
are defined for angle $d_a$ and proximal $d_p$ categories separately. Below 
is the definition of angular distance $d_a$ and proximal distance $d_p$
\begin{eqnarray}
\label{e:angular-distance}
d_a(o,c)&:=&|a_o - a_c|\\
\label{proximal-distance}
d_p(o,c)&:=&|d_o - d_c|
\end{eqnarray}
where $a_o$ is the angle to a particular object $o$ and $a_c$ is the
prototypical angle of category $c$. For example, for ``front'' the angle
 $a_c=0.0$ where 0.0 is the front side of the reference object.
$d_o$ is the distance to object $o$ and  $d_c$ is the
prototypical distance of the category $c$. 

Notice that both angular and proximal distances are always 
defined relative to a coordinate system origin. By default this is
the robot observing the world. However, robots can change
perspective to other objects, including other robots using 
their world model as basis for the transformation.

Similarity is important because it guides speakers and hearer in the 
identification referents of interactions. A speaker chooses the 
category that maximises similarity to the topic while having low similarities
with all other objects in his world model. The hearer will choose the
object as referent which has the highest similarity with the category,
he thinks the speaker is using.

\section{Acquisition of Spatial Categories}
Acquisition of a category is a two-step process. It starts
with the learner encountering a new word in a particular
communicative situation. The learner will store
the new word and the category it represents in his memory.
We call this \emph{adoption}. The information 
available to the learner in a single interaction is typically insufficient 
for a good estimate of the spatial category. The learner
will therefore integrate information from subsequent interactions
in which the new word is used. We call this process
\emph{alignment}.

\begin{figure}
\centering
\includegraphics[width=1.0\columnwidth]{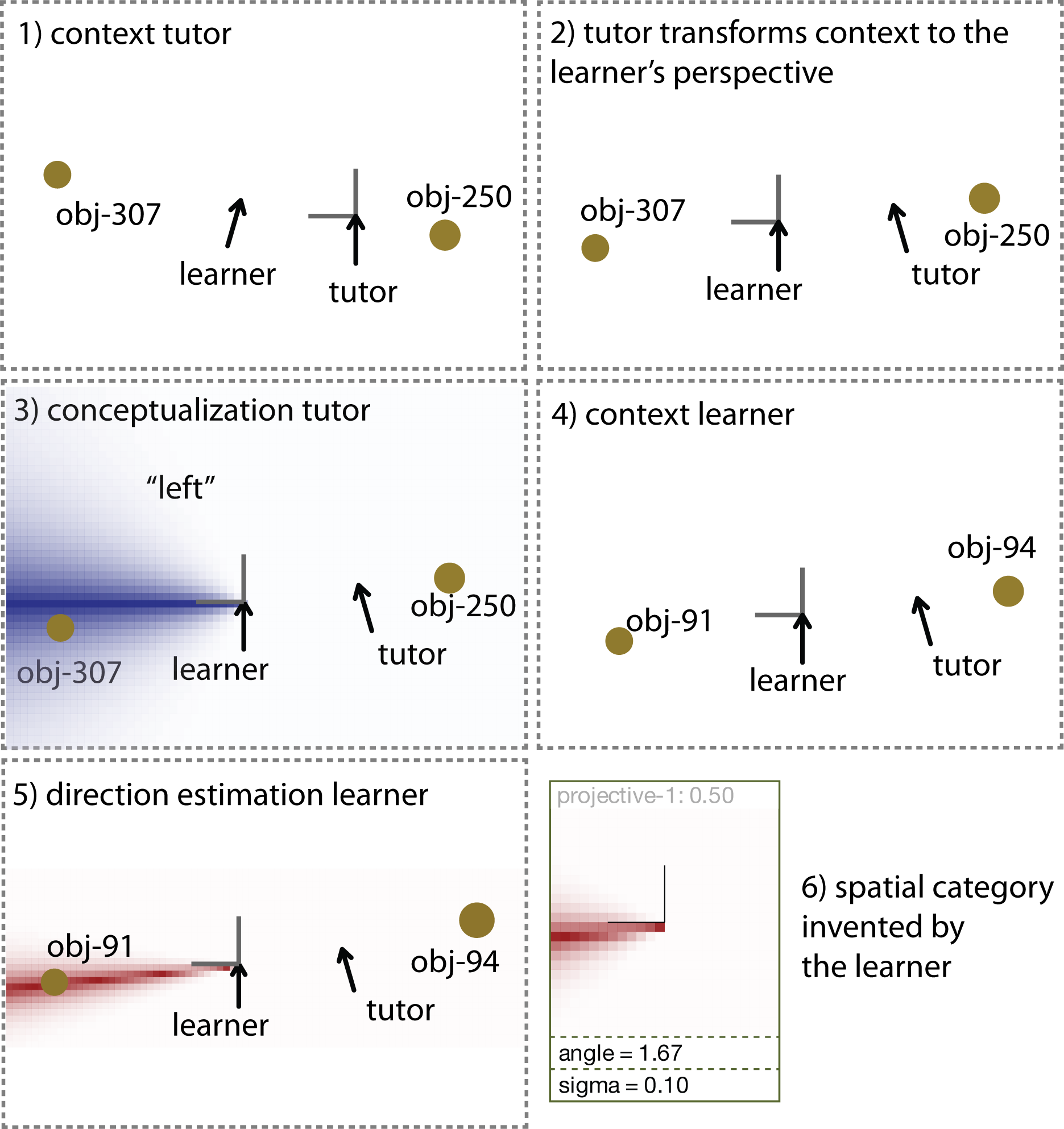}
\caption{This figure details the adoption of an unknown category label by a learner agent in
interaction with a tutor agent. The tutor who is the speaker starts by conceptualising for 
the topic object in his world model (image 1). Here, {\small\tt obj-307} ({\small\tt obj-91} in the 
learner's world model) is chosen as topic. In order to help the learner, the tutor conceptualises 
a meaning for the topic from the perspective of the learner (image 2). For this
particular topic and world model the tutor finds the category {\small\tt left}
associated with the word ``left'' to be most discriminating (image 3). The speaker then
utters the word to the learner, who himself has a particular view of the world (image 4).
When this is the first interaction ever involving the word ``left'', the learner does not know
the word and the interaction fails. However, after the speaker pointed to the topic, the
hearer can adopt the string and connect it to the newly invented projective category
{\small\tt projective-1}. The category derives its angle value from the direction to the topic object (image 5). 
The initial $\sigma$ is set to $0.1$. This is a low value that focusses the 
category around the direction of the topic object (image 6).}
\label{f:category-acquisition-projective-single-acquisition}
\end{figure}

Categories are initially adopted by a learner in a particular interaction
using the following operation, which has two parts.
The learner monitors processing and diagnoses if there
was a problem. If there was a problem, he tries to repair it.
\begin{description}
\item[\bf Hearer encounters unknown spatial term $s$]
\item[Problem:] Hearer does not know the term (step 4 fails). 
\item[Repair:] Hearer signals failure and the speaker points to the topic $t$. 
Subsequently, the hearer constructs a spatial category $c$ based on the relevant strategy 
(projective, proximal or absolute) and the topic pointed at (see Figure 
\ref{f:category-acquisition-projective-single-acquisition}). 
Additionally, the hearer invents a mapping associating $c$ with $s$. 
\end{description}

New words are always adopted in a particular interaction and in a particular context.
Angle and distance prototypes are therefore based on the particular distance and
angle of the topic of the interaction to the learner. These are never exactly
the same distance and angle measured by the tutor and more importantly never
the angle and distance of the category used by the tutor. In other words, the learner
does not have enough information to guess the category correctly or even near
correctly. Consequently, the learners require mechanisms for accumulating 
information for a particular category over time, in order, to \emph{align} to the
category used by the tutor.

Each time a category is successfully used in an interaction (by the tutor or by the learner),
the learner updates the prototype and the sigma of the category. Thereby, the learner
aligns his category representation to the tutor. For this he keeps a 
memory of past distances and angles. For example, projective 
categories are represented by prototypical angles. After each
interaction the learner updates the prototypical angle by averaging the angles 
of objects in the sample set $S$ of experiences of the category. 
The new prototypical angle $a_c$ of the category
is computed using the following formula where $a_o$ is the
angle of sample $o$.
\begin{equation}
a_c = \operatorname{atan2}\left(\frac{1}{|S|}\sum_{o\in S}\sin a_o,\frac{1}{|S|}\sum_{o\in S}\cos a_o\right)
\label{e:update-a}
\end{equation} 

The new $\sigma$ value $\sigma'$ which describes
the shape of the applicability function of the category is adapted 
using the following formula.
\begin{equation}
\sigma'_c = \sigma_c + \alpha_\sigma \cdot \left(\sigma_c - \sqrt{\frac{1}{|S|-1}\sum_{o\in S}(a_c - a_o)^2}\right)
\label{e:update-sigma-a}
\end{equation} 

This formula describes how much the new $\sigma_c$ of the category $c$ 
is pushed in the direction of the standard deviation of the sample set 
by a factor of $\alpha_\sigma \in ]0,\infty[$. 

The formula for the alignment of distance categories is the following.
\begin{equation}
d_c = \frac{1}{|S|}\sum_{o\in S} d_o
\label{e:update-d}
\end{equation} 
\begin{equation}
\sigma'_c = \sigma_c + \alpha_\sigma \cdot \left(\sigma_c - \sqrt{\frac{1}{|S|-1}\sum_{o\in S}(d_c - d_o)^2}\right)
\label{e:update-sigma-d}
\end{equation} 
Here $d_c$ is the new distance, $\sigma'_c$ the new $sigma$, $S$ the sample set, 
and $d_o$ the distance of an object in the sample set.

\subsection{Experiments and Measures}
We test the adoption operator in experiments with a population 
consisting of one tutor and one learner. The tutor is given a part of the English 
category system. The learner is only equipped 
with adoption and alignment operators. For the experiments described in 
this section, tutors and learners share the strategy used for speaking about 
reality. The tutor and student might be given just the projective
strategy or just an absolute strategy. In the latter case the tutor gets all
English absolute categories. The learner, on the other hand, gets the means to 
construe reality using the geocentric wall marker
and adoption operators that will acquire absolute categories.

Experiments are repeated 25 times\footnote{The runs are 
not directly performed on real robots but on data previously 
recorded using humanoid robots. We use a data set of over 800 scenes. 
A scene always consists of two world models. One for each robot.
Scenes differ in number and spatial configuration of objects.}. 
Each time the learner starts with an empty lexicon and no spatial 
categories. We measure the success of individual 
experimental trials using measures defined below.

\paragraph{Communicative Success} Communicative success is the most important
measure as it reflects the overall performance of the population. Every interaction
is either a success or a failure. Success is counted with $1.0$ and failure 
is counted as $0.0$. 
\paragraph{Number of Categories} This measure simply counts 
the number of categories known to the agent. It is therefore a measure of variation.
Typically one would also count 
the number of constructions (category-word mapping) an agent has. But in the acquisition experiments
described in this paper the mapping is essentially one to one. For every 
category, there is precisely one construction.
\paragraph{Interpretation similarity} This is a measure tracking how similar the 
interpretation of each word known to the tutor is to that of the learner. For this the categories
attached to the word both in the tutor and the learner are compared. We model 
projective categories by means of a direction and an applicability function width parameter 
$\sigma$. Hence, two categories are most similar when both angle and $\sigma$ are equal.
The precise formula is based on the repertoire of words $w\in W(t)$ known to the tutor  $t$
and the applicability of the category $C(t,w)$ the tutor 
associates with word $w$ to the category the learner $l$
associates with that word $C(l,w)$
\begin{equation*}
\operatorname{I}(t,l) 
=
\frac{1}{|W(t)|}\sum_{w \in W(t)} 
\operatorname{s}(C(l,w),C(t,w))
\end{equation*}
If the learner has no category associated with a particular word $w$, hence, when $C(a_{tut},w)$ 
does not find a category the applicability is $0$. If, however, the learner has a category
associated with the word then $\operatorname{s}$ is defined as follows
(e.g. for angle-based categories)
\begin{equation*}
\operatorname{s}(c,c')=e^{-\frac{1}{2}(a_c - a_c')^2\frac{2}{\sigma_c + \sigma_c'}}
\end{equation*}

\subsection{Experimental Results}

\begin{figure}
\includegraphics[width=1.0\columnwidth]{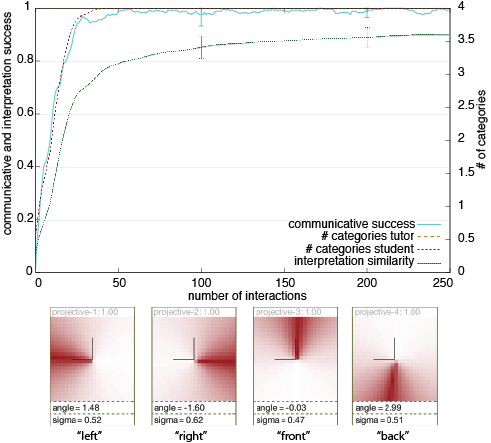}
\caption{The top figure shows the dynamics of 
acquisition experiments over 250 interactions (25 runs averaged) in which 
a learner is trying to acquire the projective language system from a tutor.
Agents quickly reach communicative success (the base line experiment of tutors communicating 
reaches approx. 98\% success for the same data set). After roughly 25 interactions, all categories and their 
corresponding strings have been adopted (the number of categories approaches 4)
In the remaining interactions the alignment operator drives the \emph{interpretation
similarity} towards 1.0 (which is the highest value and signifies total overlap between the categories
of the tutor and the learner). Interestingly, communicative success correlates
with the number of categories of the student more than it does with interpretation similarity.
This shows that agents do not need perfectly aligned categories to be able to communicate successfully.
The bottom figure shows the categories acquired by a learner in one 
particular population of an acquisition experiment and to which strings they are linked. 
The resulting categories are very similar to the projective categories given to the tutor.}
\label{f:category-acquisition-projective-results+categories}
\end{figure}

Figure \ref{f:category-acquisition-projective-results+categories} shows aggregated
dynamics of 25 experimental runs testing the acquisition of projective categories. 
The learner quickly reaches communicative success which means he can act
successfully as hearer and speaker after 25 interactions. He learns all categories
from the tutor (in this case the four spatial categories).

\begin{figure}
\centering
\includegraphics[width=1.0\columnwidth]{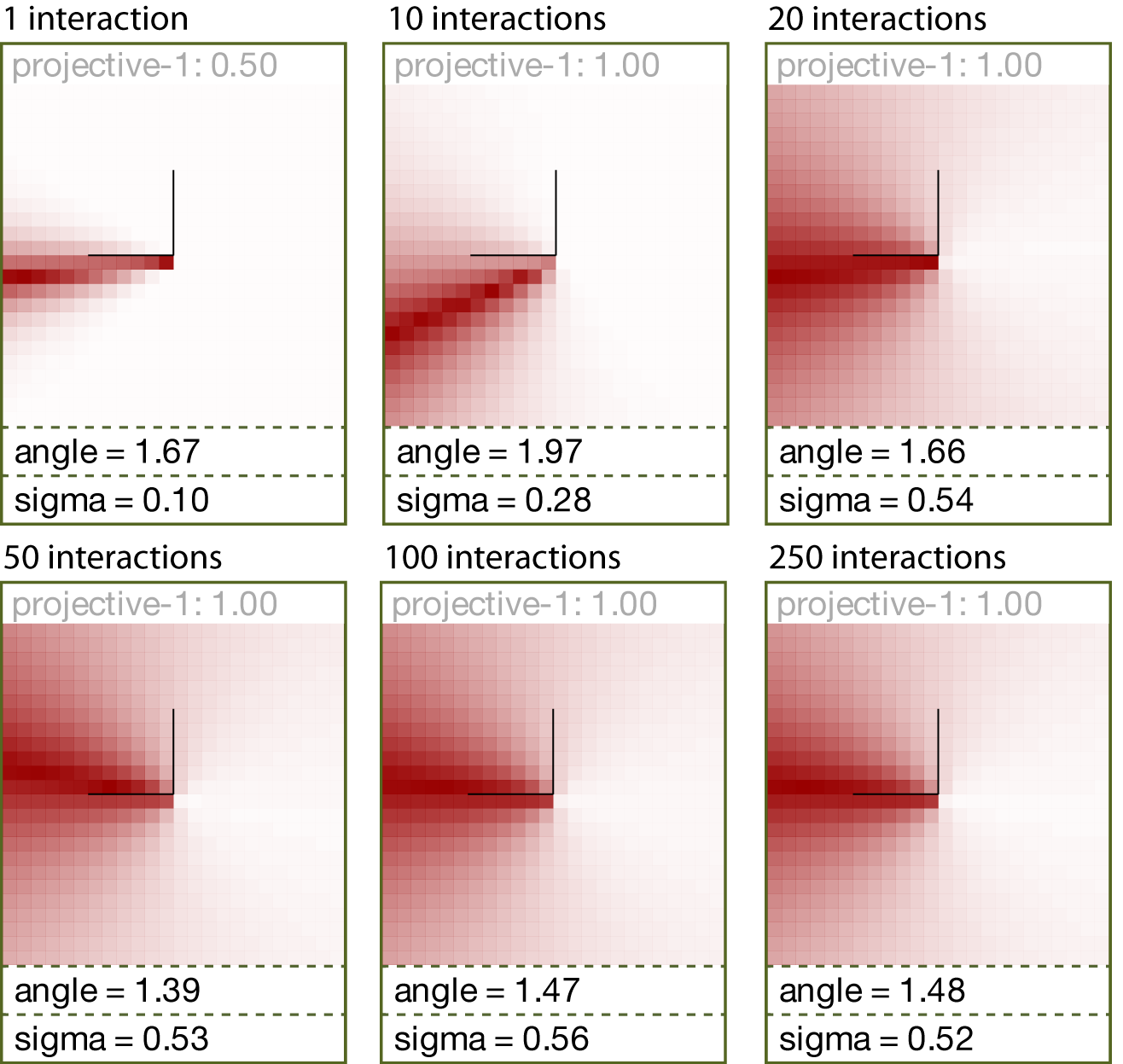}
\caption{Development of the projective category whose initial adoption is depicted in 
Figure \ref{f:category-acquisition-projective-single-acquisition} over many interactions 
(after 1, 20, 50, 100 and 200 interactions). In the beginning the width of the category is narrow (small $\sigma$). Gradually 
its direction approaches the direction of the target category {\small\tt left} and so does its $\sigma$ 
(the target category's $\sigma$ is $0.4$).}
\label{f:category-acquisition-projective-development-over-time}
\end{figure}

Figure \ref{f:category-acquisition-projective-development-over-time} shows how 
the alignment operator makes categories 
evolve over time (same category as in Figure 
\ref{f:category-acquisition-projective-single-acquisition}). The 
categories of the learner become aligned with the operators of the tutor over time.
This is also apparent from the dynamics of the  interpretation similarity measure.

Naturally, alignment and
adoption operators have a number of parameters, for instance
how many samples to consider, how eager to update the $\sigma$ component using
$\alpha_\sigma$ and so on and so forth. We have tested the impact of these parameters
by doing many repeated experiments with different parameter settings (results not shown
for space reasons). We can show that these parameters are quite robust and small 
changes do not affect the overall performance of the system.

\section{Simultaneous Acquisition of Various Spatial Systems}

In the experiments just described, there is only a single category system at work.
Learners are acquiring either a projective, an absolute, or a proximal system.
Obviously this is a limitation of the model and the experiments. Human students 
of English are learning multiple systems more or less at the same time. At least 
students are confronted with different systems at the same time. 

Suppose a tutor is given the three strategies: projective, proximal and absolute. 
Suppose further that the learner agent in such conditions hears a new term, for 
example, ``near'' from the tutor. The learner faces a problem. He has no a priori 
means of knowing whether this term is part of a projective, proximal or absolute strategy.
What is necessary to allow artificial learners to cope with a situation where there
is uncertainty about the strategy used by the tutor? One possibility
often discussed in the literature on language is to use discrimination as a means to
decide between various conceptualisation strategies. We know from
discourse theory that language users are maximally cooperative and choose 
most discriminating strategies in verbalisation \cite{grice1975logic,sperber1986relevance}.
These facts have also been confirmed for spatial language \cite{herskovits1986language}.
A way for a student to decide which conceptualisation
strategy underlies an unknown term is to see which strategy 
is most discriminating.

\begin{figure}
\centering
\includegraphics[width=1.0\columnwidth]{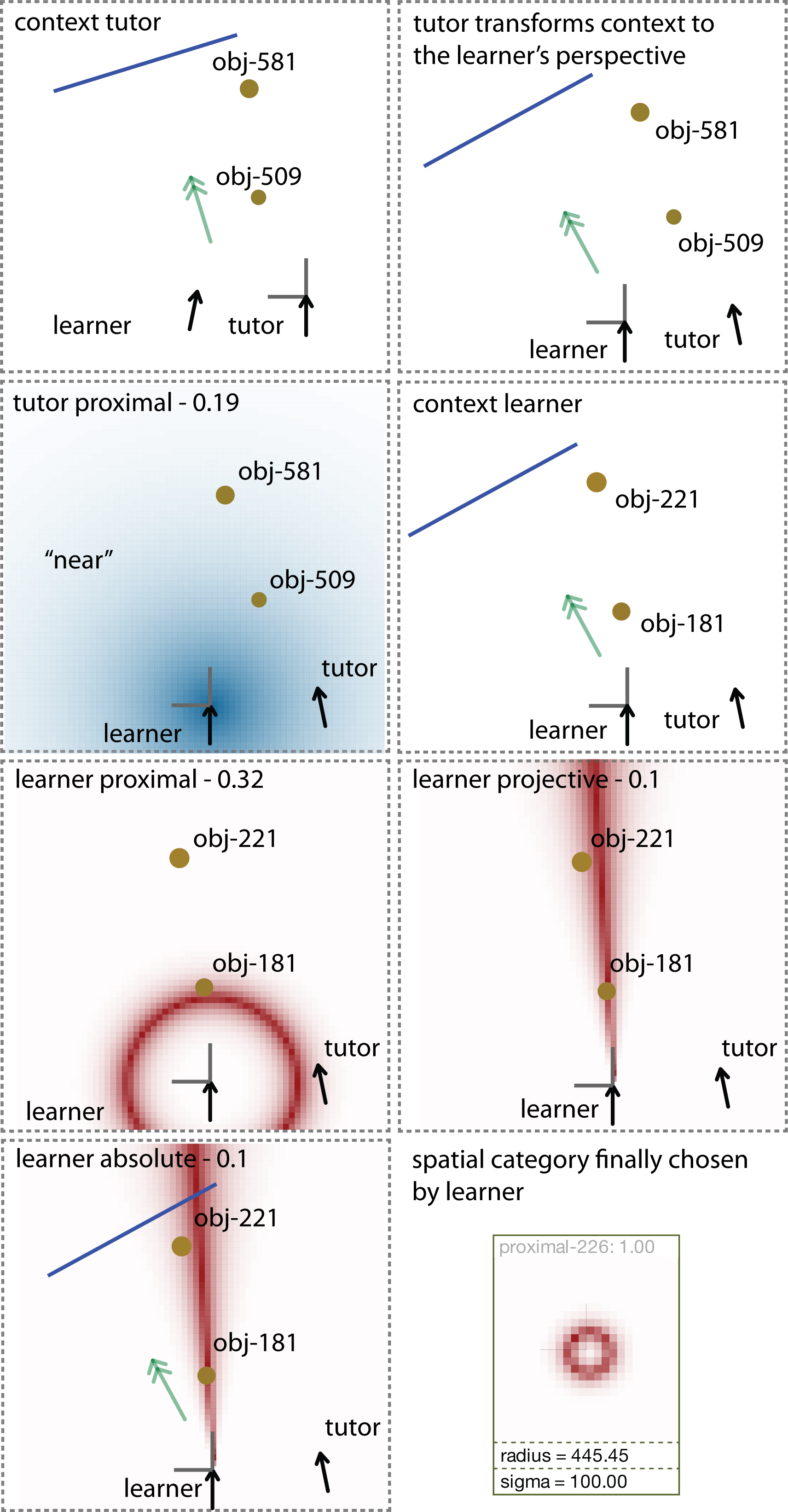}
\caption{Inference by the learner in re-conceptualisation 
(after receiving pointing from the speaker) as to which 
category type was used by the tutor (speaker). 
The steps are the same as for proximal, projective and 
absolute category adoption. However, instead of 
just adopting one category, the learner will
build three categories for the topic object.
Each of these categories has a particular discrimination
score. The learner chooses to store the category
with the highest discrimination score.
Here the invented proximal category wins (score 0.32).
It is the category that will be linked to the word ``near''.}
\label{f:category-acquisition-absolute+proximal+projective-single-acquisition}
\end{figure}

Figure \ref{f:category-acquisition-absolute+proximal+projective-single-acquisition} 
gives a graphical explanation of the process. Upon hearing a new 
word, the learner invents for all conceptualisation strategies. Each of these inventions
will lead to a potential candidate category. The learner tries each of these categories
in his current world model and computes a discrimination score. The learner only 
keeps the category which is most discriminating.  Discriminative power of candidate 
categories is ranked via scores. The score is based on the similarity of a particular 
category with the topic object and the difference of that with the similarity score of 
other (distractor) objects in the world model. Agents choose the candidate category with 
the highest score.

\begin{eqnarray}
\operatorname{disc}(c,t)=\operatorname{sim}(c,t) - \max_{o \in C/t}(\operatorname{sim}(c,o))
\end{eqnarray}
where $c$ is a candidate category and $t$ is the topic object.

\begin{figure}
\includegraphics[width=0.9\columnwidth]{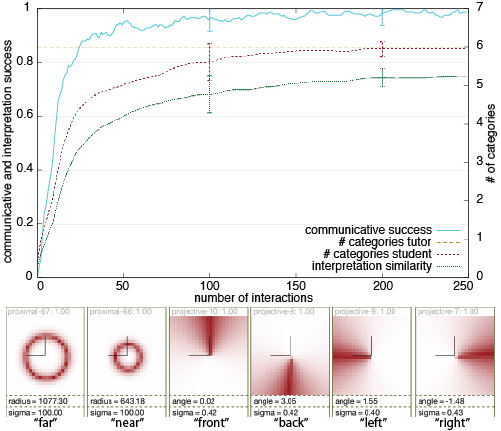}
\caption{Results of acquisition experiments using inference. 
The tutor is equipped with proximal and projective 
categories.}
\label{category-acquisition-projective+proximal-results+categories}
\end{figure}

Figure \ref{category-acquisition-projective+proximal-results+categories} shows results
for the simultaneous acquisition of proximal and projective categories. The learner quickly 
acquires the target system (near, far, north, south, east, west). Similar results can be 
obtained for the simultaneous acquisition of proximal and absolute categories. 

Discriminative power is not a perfect ``guessing'' mechanism. The reason
is twofold. First, the experiments are grounded. Robots frequently make different 
distance and angle estimations of the same object. This can make the learner
guess the wrong strategy because discrimination scores rely on perceived angles and
distances. Second, tutors have a fixed category system which has very particular 
distances and angles encoded in category prototypes. For example, the {\small\tt left} 
prototype is exactly at 90 degrees. The learner though will guess a 
category based on the distance and angle to the topic which is never directly aligned
to 90 degree angles. Consequently, similarity and therefore discrimination score
are different functions for tutor and learner.

Obviously, for discrimination to work, there needs to be sufficient
discriminative difference between strategies. Otherwise, learners 
have no means of deciding between different strategies in particular
interactions. This limitation already plays out with the spatial systems
discussed in this paper. There are two angle-based strategies: projective
and absolute. Here discrimination fails because as can be seen in Figure 
\ref{f:category-acquisition-absolute+proximal+projective-single-acquisition}
both strategies have the same discrimation score (0.1). As a consequence
experiments where learners have to acquire the projective and absolute system
at the same time additional mechanisms are needed. 

Such mechanisms can be biases or other information that clearly
distinguish between system. For instance, absolute and projective could be
treated differently in the grammar (not the case for English). 
Also, the English absolute system is much less frequently used for referring 
to objects in the immediate surroundings of interlocutors, and it is tied to particular 
measurement devices such as compass or sun position estimation. 
In other words, to learn the absolute system requires to understand
particular human technology. Something not modelled here.

\section{Conclusion}
The most important result of this paper is to show how
to go from single strategy acquisition to a system that can 
acquire multiple category systems at the same time.
This is an important result and it shows the potential for scale 
up, particularly with respect to grounded systems that learn from 
increasingly unconstrained data. 

This paper detailed a number of experiments exploring the acquisition 
of spatial language by grounded, artificial agents.
Through the proposed learning operators (adoption and alignment), learners were enabled
to successfully acquire different target systems. Ongoing and future research is building 
on the results presented in this paper and extends them to grammar and complex 
semantic structure acquisition in interactive scenarios.

\section*{Acknowledgment}
I would like to thank Masahiro Fujita and his team
at Sony Corporation for providing the robots and technical
support. Luc Steels and Martin Loetzsch helped with the 
implementation of the robotic setup. 
Funding for this research was provided by the EU projects FP6 
ECAgents and FP7 ALEAR.

\bibliographystyle{IEEEtran}
\bibliography{papers}

\end{document}